\documentclass{article}
\usepackage{caption}
\usepackage{diagbox}
\usepackage{amsmath}
\usepackage{spconf,amsmath,graphicx}
\usepackage[table]{xcolor}
\usepackage{times}
\usepackage{enumitem}

\setlist{nosep, leftmargin=14pt}
\usepackage{mwe} 
\usepackage{graphicx}
\usepackage{caption}
\usepackage{cite}

\usepackage{multirow}
\title{FMIR, a foundation model-based Image Registration Framework for Robust Image Registration}
\name{Fengting Zhang$^{1}$, Yue He$^{1}$, Qinghao Liu$^{1}$, Yaonan Wang$^{1}$, Xiang Chen$^{1,}$$^{\dagger}$\thanks{$^{\dagger}$Corresponding author: xiangc@hnu.edu.cn(Xiang Chen)}, Hang Zhang$^{2}$}

\address{$^{1}$ School of Artificial Intelligence and Robotics, Hunan University, China\\
$^{2}$ Department of Electrical and Computer Engineering, Cornell University, USA}

\begin{document}
%
\maketitle

\begin{abstract}
Deep learning has revolutionized medical image registration by achieving unprecedented speeds, yet its clinical application is hindered by a limited ability to generalize beyond the training domain, a critical weakness given the typically small scale of medical datasets.
In this paper, we introduce FMIR, a foundation model-based registration framework that overcomes this limitation.
Combining a foundation model-based feature encoder for extracting anatomical structures with a general registration head, and trained with a channel regularization strategy on just a single dataset, FMIR achieves state-of-the-art(SOTA) in-domain performance while maintaining robust registration on out-of-domain images.
Our approach demonstrates a viable path toward building generalizable medical imaging foundation models with limited resources. The code is available at https://github.com/Monday0328/FMIR.git.

\end{abstract}
\section{Introduction}
\label{sec:intro}

Image registration, a cornerstone of medical image analysis, aims to spatially align two or more images acquired at different times, with different modalities, or from different subjects by estimating optimal transformation parameters or deformation fields. This process can be broadly categorized into affine registration and deformable registration, and is a technique critical to numerous downstream applications, including image fusion, motion tracking, disease diagnosis, and surgical navigation \cite{chen2021deepSurvey}.


Traditional registration methods \cite{avants2011reproducible} typically iteratively optimize deformation fields/parameters under the guidance of objective functions combining dissimilarity metrics (e.g., Mean Square Error (MSE) or Normalized Cross-Correlation (NCC)) and smoothness regularization. Although these methods achieve accurate registration across various registration tasks, their iterative nature during inference renders the process computationally expensive and time-consuming\cite{balakrishnan2019voxelmorph,chen2022transmorph}.


Deep learning-based methods have revolutionized this task. By training on pre-defined datasets, these networks achieve registration speeds that are orders of magnitude faster (requiring only a single forward pass, generally $<1$s) while delivering accuracy comparable to conventional iterative methods. A central development for deep learning-based registration has been receptive field expansion: early U-Net architectures\cite{balakrishnan2019voxelmorph, chen2021deepDiscontinuity} exhibited limited contextual awareness; subsequent improvements incorporated transformer modules\cite{chen2022transmorph}  and large-kernel convolutions\cite{jia2022u} for enhanced global reasoning; pyramid architectures\cite{memwarp2024, wang_RDP, chen2025encoder} and recursive frameworks\cite{zhao2019recursive} further advanced large deformation handling through progressive estimation.
Despite their impressive performance on in-distribution test data, these learning-based methods share a critical limitation: an inherent assumption that training and test data follow the same distribution. This assumption often leads to implausible registrations and poorer generalization to unseen domains compared to traditional optimization-based techniques.

Recent advances in foundation models have demonstrated remarkable capabilities across various domains, with models like the Segment Anything Model (SAM) \cite{mazurowski2023segment} and DINOv3 \cite{simeoni2025dinov3} (abbreviated as DINO hereafter) showing exceptional performance in natural image segmentation tasks. Although originally developed for 2D natural images, their encoders pre-trained in large-scale data learn semantically rich representations that transfer effectively to medical imaging tasks, including 2D and 3D segmentation \cite{MedSAM}. However, while foundation models have been widely adopted in medical image segmentation, their application to medical image registration remains largely unexplored, with limited attempts such as uniGradICON \cite{uniGradICON2024MICCAI} demonstrating robust performance at the cost of complex architectures and prolonged inference times. 
Although foundation models provide a promising alternative to traditional deep learning methods by enabling zero-shot inference without dataset-specific retraining, their practical deployment still faces challenges. The substantial computational resources and large-scale datasets necessary for training these models from scratch present a significant trade-off, posing practical hurdles for their broader adoption.

To address this limitation and enable efficient yet robust registration under constrained resources, we propose a novel foundation model-based image registration framework (FMIR). Our method comprises a foundation model-based feature encoder for extracting semantically rich representations and a general registration head for deformation field prediction. Evaluated on two distinct datasets, our approach demonstrates that a model trained on only a single dataset can still achieve robust performance even on significantly different image domains.
The main contributions of this work are summarized as follows:
\begin{itemize}
\item We introduce a novel foundation model-based registration framework that, once trained on a single dataset, maintains strong performance on out-of-domain images, offering a promising pathway toward practical registration foundation models.
\item We design a general registration head that captures motion correlations across multi-scale, multi-shape pyramid features extracted by the foundation encoder, enabling flexible integration with various foundation encoders.
\item We propose a channel regularization strategy that encourages the model to learn essential structural correlations rather than overfitting to dataset-specific priors, greatly enhancing cross-domain generalization.
\end{itemize}

\section{Method}
\label{sec:method}

\subsection{Overall Framework}
\begin{figure*}[htbp]
    \centering
    \includegraphics[width=1.0\linewidth]{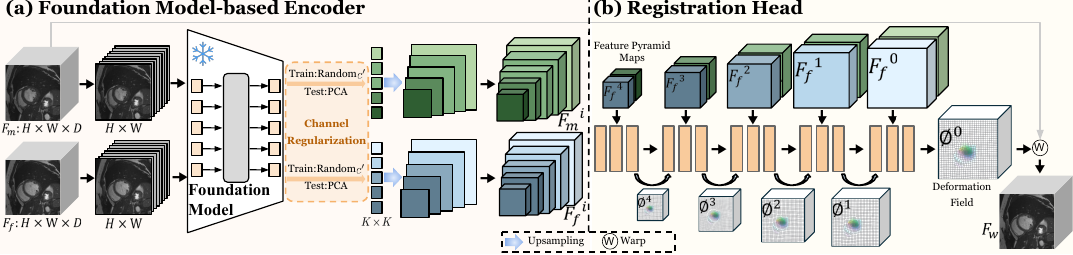}
    \caption{
    The schema of our FMIR: a Foundation Model-based Encoder and a Registration Head.
    }
    \label{fig:fmir}
    \vspace{-2ex}
\end{figure*}

Image registration aims to predict a deformation field $\textbf{u}$ that establishes dense point correspondence between a moving image $\textbf{I}_m$ and a fixed image $\textbf{I}_f$. This defines a deformation function $\boldsymbol{\phi}(x) = x + \mathbf{u}(x)$, which transforms a point $x$ in $\textbf{I}_m$ to its corresponding location in $\textbf{I}_f$. As illustrated in Fig. \ref{fig:fmir}, our FMIR consists of two core components: 1) a foundation model-based feature encoder that extracts domain-invariant and organ-invariant features $\textbf{F}_m$ and $\textbf{F}_f$ from $\textbf{I}_m$ and $\textbf{I}_f$, respectively; and 2) a registration head to learns the mapping from these features to the final deformation field $\textbf{u}$.

\subsection{Foundation Model-based Encoder}

This module leverages pre-trained 2D foundation models to extract domain-invariant features, enhancing registration robustness against domain shift. While trained on natural images, their captured visual primitives demonstrate strong cross-domain generalization, enabling semantically meaningful feature extraction without medical imaging priors.
To adapt these 2D models for 3D medical data, we design a slice-based processing pipeline. Each volume $R \in H\times W \times D$ is decomposed into $D$ slices of size $H\times W$. For models requiring square inputs (e.g., SAM, DINO), slices are padded to $K\times K$ when necessary. Each slice is processed by the encoder of a frozen foundation model, producing feature maps of size $c \times \frac{K}{16}\times \frac{K}{16}$ (where DINO: $c=768$, SAM: $c=256$).

A channel regularization then is applied to reduce feature dimensionality into a unified $c'=256$ (in this paper, we train FMIR with DINO), mitigating computational costs and redundant features. The processed 2D features are reconstructed to 3D volume $c' \times H\times W\times D$ through reassembly, upsampling, and cropping. While this slice-wise approach naturally suppresses 3D-specific biases, we apply a 3-layer 3D convolutional block to recover local volumetric context and further compress channels to $n<<c'$. The encoder maintains flexible backbone compatibility, requiring only minimal adjustments to channel dimensions when integrating new foundation models.


\subsection{Registration Head}
Following the feature extraction stage, the registration head estimates the deformation field using a multi-scale pyramid structure to effectively handle large deformations. The learned features of the moving and fixed images, $\textbf{F}^i_m$ and $\textbf{F}^i_f$, are first downsampled via trilinear interpolation to construct a five-level feature pyramid, denoted as $\{\textbf{F}^i_m,\textbf{F}^i_f\}^4_0$. At each level $i$, the corresponding feature pair is processed by a three-layer convolutional block to predict a residual deformation field $\textbf{u}^i$. The final deformation field is reconstructed progressively from the coarsest to the finest scale: at each level, the upsampled deformation field from the previous (coarser) level is composed with the current level's predicted field, effectively breaking down a large displacement into a sequence of smaller, more manageable deformations. This coarse-to-fine strategy significantly improves the stability and accuracy of the registration. Once trained, the same registration head can be directly applied to features from any foundation model backbone, ensuring broad compatibility. 
For models with differing output channel dimensions, simple techniques such as random channel selection or principal component analysis (PCA) can be employed to standardize the feature channels, maintaining the general applicability of our FMIR framework without structural changes.

\subsection{Channel Regularization (CR)}

To fully leverage the modality-invariant characteristics of foundation model features and encourage the registration head to learn from fundamental visual correlations rather than dataset-specific priors, we introduce a channel regularization strategy. During training, we randomly select a subset of $c'$ channels from the full feature set for each forward pass, acting as a form of channel `dropout'. This forces the network to avoid over-relying on any fixed set of feature channels and promotes robust, generalized learning. During inference, we replace stochastic channel selection with a deterministic channel reduction using PCA, projecting the features into a $c'$-dimensional space that preserves the most salient information. This regularization mechanism explicitly differentiates the features used during training from those in testing. Therefore, it effectively suppresses the model's tendency to rely on spurious, data-specific patterns and instead directs it to focus on capturing the essential correlations between moving and fixed image features. 

\subsection{Loss Function}
To train FMIR, we employ Normalized Cross Correlation (NCC) \cite{balakrishnan2019voxelmorph, chen2024textscf} as the similarity metric, alongside a diffusion-based smoothness loss $L_{smooth}$ (following \cite{balakrishnan2018unsupervised}) applied to the deformation field's gradients to ensure spatial coherence. For weakly-supervised learning, the Dice loss $L_{Dice}$ is added as an auxiliary similarity loss. The complete loss function is:
\begin{equation}
L = \lambda_0 L_{NCC} + \lambda_1 L_{Dice}
+ \lambda_2 L_{smooth}, 
\end{equation}
where $\lambda_0$-$\lambda_2$ are hyper-parameters defined empirically.


\section{Results}

\subsection{Datasets and Implementation Details}
\textbf{Datasets.}
We evaluate FMIR on two public benchmarks: the ACDC cardiac Magnetic Resonance (MR) dataset \cite{bernard2018deep} and the Abdomen Computed Tomography (CT) dataset \cite{xu2016evaluation} from Learn2Reg 2020. For intra-subject cardiac registration on ACDC, we use 85 subjects for training, 15 for validation, and 50 for testing. Each subject provides an end-diastole (ED) and end-systole (ES) frame with expert segmentations, yielding 170/30/100 (train/val/test) registration pairs from both ED-to-ES and ES-to-ED directions. All MR volumes were preprocessed to a size of $128 \times 128 \times 16$ with a voxel spacing of $1.8 \times 1.8 \times 10 mm^3$. For inter-subject registration, the Abdomen CT dataset comprises 30 scans with 13 annotated organs. Images were resampled to a $192\times160\times256$ at 2 mm isotropic resolution and split into 20 training, 3 validation, and 7 test scans. This results in 380, 6, and 42 registration pairs for training, validation, and testing, respectively.

\textbf{Implementation Details.}
In our experiments, all models were developed using PyTorch on an A6000 GPU machine.  We utilize Adam optimizer for network training, with an initial learning rate of $1e^{-4}$ and a polynomial learning rate scheduler with a decay rate of 0.9. $\lambda_0$-$\lambda_2$ are all set as 1 (for unsupervised training, $\lambda_1=0$). We set $K=512$ and $n=32$ in FMIR. To train our FMIR, we only utilize DINO ViT-B model, but SAM is also utilized in the inference to validate the generalization of our framework.


\textbf{Comparison Methods and Metrics.}
We conduct a comprehensive evaluation of our FMIR framework against SOTA learning-based registration models. All competing methods are sufficiently trained on the same datasets as FMIR, with the exception of uniGradICON~\cite{uniGradICON2024MICCAI}, which, given its foundation model nature, is directly applied without retraining. Following~\cite{balakrishnan2019voxelmorph, hering2022learn2reg,wang_RDP}, we assess registration quality using the Dice Similarity Coefficient (Dice) and 95\% Hausdorff Distance (HD95) to quantify anatomical alignment accuracy. The smoothness of deformation fields is evaluated using the standard deviation of the logarithm of the Jacobian determinant (SDlogJ). Furthermore, the registration time is also reported.
\subsection{Experiments and Analysis}

\begin{table}[!h]
\begin{center}
\caption{
Quantitative comparison on the cardiac ACDC dataset. 
Best-performing metrics are highlighted in bold. 
Symbols indicate direction: $\uparrow$ for higher is better, $\downarrow$ for lower is better. 
`Initial' refers to baseline results before registration.
`Time' is the average inference time for one registration pair.
`-CR' denotes removing the channel regularization on FMIR and training in an unsupervised manner.}
\label{tab:ACDC}
\scriptsize
\begin{tabular}{ lccccrc }

\hline
\rowcolor{lightgray}
Model & Dice (\%) $\uparrow$ & HD95 (mm) $\downarrow$ & SDlogJ $\downarrow$ & Time (s) $\downarrow$ \\ 
\hline
Initial & 58.14 & 11.95 & - &- \\
\hline
VoxelMorph \cite{balakrishnan2018unsupervised} & 75.26 & 9.33 & \textbf{0.044} & \textbf{0.18} \\
TransMorph \cite{chen2022transmorph} & 74.97 & 9.44 & 0.045 & 0.26 \\
LKU-Net \cite{jia2022u} & 76.53 & 9.13 & 0.049 & 0.22 \\
\hline
CorrMLP \cite{meng2024correlation} & 77.31 & 9.00 & 0.056 &  0.28 \\
MemWarp \cite{memwarp2024} & 76.74 & 9.67 & 0.108 & 0.58 \\
RDP \cite{wang_RDP} & 78.06 & 9.02 & 0.076 & 0.36 \\
\hline
uniGradICON \cite{uniGradICON2024MICCAI} & 78.89 & 9.05 & 0.049 & 4.95 \\
\hline
\textbf{FMIR}& 79.82& 9.07 & 0.049 & 0.62\\
\textbf{FMIR(SAM)}& \textbf{79.72}& 9.06 & 0.046& 1.79\\
\textbf{FMIR(-CR)}& 79.43& \textbf{8.57} & 0.045& 0.61\\
\hline
\end{tabular}
\end{center}
\vspace{-2ex}
\end{table}

\textbf{Registration on In-domain Data.}
We first evaluate the in-domain unsupervised performance of FMIR on intra-subject cardiac MR registration. As summarized in Table \ref{tab:ACDC}, pyramid-based methods (CorrMLP, MemWarp, and RDP) generally outperform U-Net-based architectures (VoxelMorph, TransMorph, and LKU-Net). The foundation model uniGradICON outperforms the previous methods but requires longer inference time due to its complex pre-processing and post-processing. In comparison, our FMIR surpasses uniGradICON, with significantly less time (note that, feature extraction time is also included for FMIR). This demonstrates FMIR's ability to harness foundation model features for enhanced performance while maintaining high efficiency.

\begin{table}[h]
\centering
\caption{
Cross-dataset validation of FMIR, comparing unsupervised(UN),weakly-supervised(WS),and unsupervised without channel regularization(-CR)training.
}
\footnotesize
\begin{tabular}{cccccc}
\hline

\rowcolor{lightgray}
\multicolumn{2}{c}{\diagbox[width=2.8cm]{Train Set}{Test Set}}& \multicolumn{2}{c}{ACDC(Dice(\%))$\uparrow$} & \multicolumn{2}{c}{Abdomen(Dice(\%))$\uparrow$}\\ 
\rowcolor{lightgray}
\multicolumn{2}{c}{} & DINO & SAM & DINO & SAM\\ \hline
Initial & & \multicolumn{2}{c}{58.14} & \multicolumn{2}{c}{30.86}\\ \hline
\multirow{3}{*}{ACDC} &UN& \underline{79.82} & \underline{79.72} & 40.16 & 40.20\\ 
 & WS & \textbf{87.15} & \textbf{87.19} & 37.12 & 36.77\\
 & -CR & 79.43 & 69.18 & 38.90 & 34.18 \\\hline
\multirow{3}{*}{Abdomen} & UN & 73.34 & 73.52 & 49.11 & 49.04 \\ 
 & WS & 70.85 & 71.33 & \textbf{56.79} & \textbf{57.50}\\
  & -CR & 73.14 & 70.28 & 50.50 & 44.24\\\hline
\multirow{3}{*}{Hybrid} & UN & 78.89  & 78.88 & \underline{50.87} & \underline{50.99} \\ 
 & WS & 76.24 & 74.35 & 44.52 & 42.73\\  
 & -CR & 79.04 & 72.75 & 50.92 & 45.06\\\hline
\end{tabular}
\label{tab:cross-dino}
\end{table}

\textbf{Registration on Out-of-domain Data.}
To evaluate the out-of-domain generalization capability of our FMIR framework, we construct three training configurations: using only the ACDC dataset, only the abdomen CT dataset, and a hybrid combination of both (denoted as `Hybrid'). For each configuration, we train FMIR under both unsupervised (UN) and weakly-supervised (WS) settings. As shown in Table~\ref{tab:cross-dino}, models trained with weak supervision achieve superior performance on in-domain test data. However, when evaluated on out-of-domain data, their unsupervised counterparts demonstrate stronger generalization. We attribute this phenomenon to the fact that weakly-supervised training introduces stronger domain-specific priors, which may hinder adaptability to unseen data distributions. Furthermore, models trained on the hybrid dataset underperform those trained on a single dedicated dataset, as the substantial differences between cardiac and abdominal motion patterns force the network to compromise between conflicting objectives. Notably, even when trained solely on abdominal data, our FMIR achieves registration accuracy on cardiac images comparable to TransMorph trained directly on the ACDC dataset, highlighting the enhanced robustness and domain invariance of our approach.

\textbf{Analysis of Channel Regularization.}
To validate the effectiveness of our channel regularization strategy, we conduct an ablation study by replacing it with a more conventional approach: applying PCA for channel reduction during both training and inference (denoted as `-CR'). As shown in Table \ref{tab:cross-dino}, removing our channel regularization (`-CR') results in comparable in-domain performance but comes at a great cost to out-of-domain generalization. This is especially evident when integrating with SAM, where performance severely degrades without our CR strategy. To our analysis, using PCA on both training and testing induces over-reliance on dataset statistics, leading to early convergence on superficial features. Our approach, however, explicitly forces the model to capture essential structural correspondences between features. The significant performance decline without CR strategy highlights its critical role in discarding data-specific priors and learning robust, generalizable features for cross-domain alignment.


\begin{figure}[ht]
\begin{center}
\includegraphics[width=\linewidth]{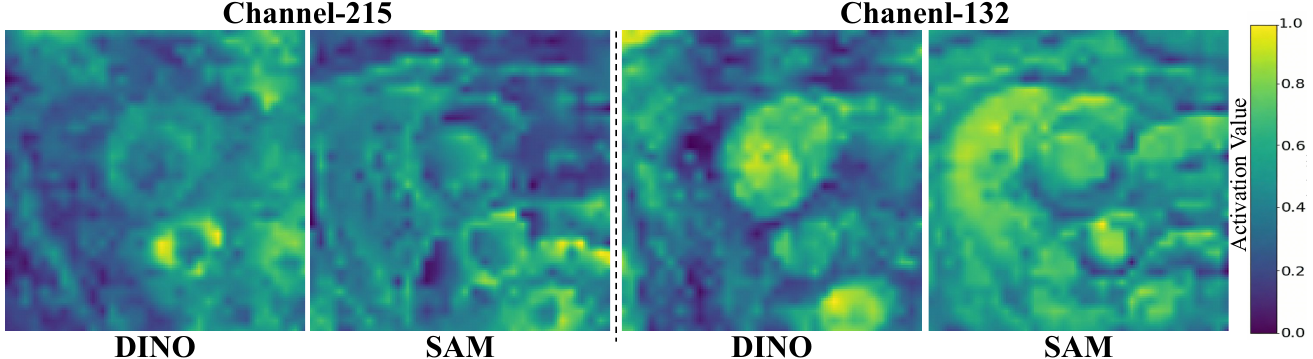}
    \caption{
    Channel visualization on the DINO and SAM features.
    }
    \label{fig:channel_visual}
    \vspace{-2ex}
    \end{center}
\end{figure}

\textbf{SAM vs DINO.}
We further demonstrate the flexibility of our FMIR framework through a plug-and-play replacement of the foundation model within the encoder. Although trained on DINO features, FMIR can directly operate on SAM-encoded features without any retraining. As shown in Table \ref{tab:cross-dino}, when integrated with SAM, FMIR achieves registration accuracy comparable to its DINO-based counterpart, even showing a marginal improvement in some scenarios. This interoperability is notable given the significant differences in the feature structures highlighted by DINO and SAM (see Fig. \ref{fig:channel_visual}). The fact that these distinct feature sets lead to comparable performance strongly demonstrates that our registration head has learned a generalizable correspondence mapping, rather than relying on feature priors specific to any particular dataset. This confirms its ability to consistently achieve robust performance across both in-domain and out-of-domain data.



\section{Discussion and Conclusion}


This paper presents FMIR, a robust registration framework that leverages foundation models pre-trained on 2D natural images to achieve SOTA in-domain performance while maintaining strong generalization on out-of-domain data. By integrating a foundation model-based encoder with a general registration head, FMIR extracts domain-invariant features for accurate deformation estimation. A channel regularization strategy is further incorporated to effectively suppress dataset-specific biases. Extensive evaluations across multiple tasks and backbones validate its registration performance and generalization capability. Given the finite anatomical variation in human organs, developing a universal medical image registration model is increasingly feasible. FMIR establishes an adaptable baseline toward this goal, with future work focusing on integrating more advanced foundation model encoders and expanding multi-anatomy datasets to enhance robustness and universality.

\section{Compliance with ethical standards}

This research utilized exclusively publicly available datasets. No new experiments involving human participants or animals were conducted by the authors. Therefore, ethical approval from an institutional review board was not required for this study. 

\section{Acknowledgments}
This work was supported in part by the National Natural Science Foundation of China under grant 62503161, and in part by the science and technology innovation Program of Hunan Province under grant 2025RC3069, and in part by the Natural Science Foundation of Hunan Province under Grant 2025JJ60389.

\bibliographystyle{IEEEbib}
\bibliography{fmir}

\end{document}